\documentclass{article}
\usepackage{colm2024_conference}

\usepackage{amsmath,amssymb}
\usepackage{booktabs}
\usepackage{array}
\usepackage{algorithm}
\usepackage{algpseudocode}

\usepackage{graphicx}
\usepackage{enumitem}
\usepackage{xspace}

\newcommand{\dataname}{UserSWE\xspace}
\newcommand{\nsessions}{20{,}859\xspace}

\newcommand{\fix}[1]{\emph{#1}}

\title{\dataname: Evaluating Coding Agents \\ in Multi-Turn Conversations with Dynamic Users}

\author{\textbf{Zhan-Yu Wu}$^{1,\ast}$, \textbf{Ji Luo}$^{1,\ast}$, \textbf{Alyssa Wu}$^{1}$, \textbf{Junjielong Xu}$^{1}$,\\
\textbf{Richeng Xuan}$^{1}$, \textbf{Zhichao Hu}$^{1,\dagger}$, \textbf{Chongming Gao}, \textbf{Xinting Hu}$^{\dagger}$\\[0.45em]
{\normalsize\normalfont $^{1}$Tencent Hunyuan}}

\begin{document}
\maketitle
\let\oldthefootnote\thefootnote
\let\thefootnote\relax
\footnotetext{$^{*}$Equal contribution.\hspace{1.2em}$^{\dagger}$Correspondence to qsenre@gmail.com, cheeseluo@tencent.com.}
\let\thefootnote\oldthefootnote

\begin{abstract}
As coding agents become more capable, developers increasingly use them to fix bugs
through multi-turn interactions. Existing benchmarks do not adequately cover this
scenario. Most present a complete issue report in a single turn without interaction. 
Their evaluation can produce both false positives and false negatives, accepting
incorrect patches or rejecting correct ones. We introduce \dataname for evaluating
coding agents' ability to fix repository-level bugs through multi-turn interactions
with dynamic users and execution-grounded verdicts at every turn. Its construction
pipeline turns repository history and existing single-turn benchmarks into multi-turn
tasks. At each turn, information pools constrain what the simulated user can reveal, while a persona chain sampled from a transition model calibrated on \nsessions development sessions governs how it speaks. Tests run after every turn, and a pair
of agents from different model families writes and gates each verdict, using test
results as primary evidence and model judgment only for aspects the tests cannot
resolve. We evaluate eleven agents on 72 tasks from 42 repositories and find that
agent rankings change with the interaction budget. Claude Fable 5.1 resolves the
most tasks in one turn, GPT-6-Astra resolves the most by the eighth turn, and Claude
Opus 5 leads at most intermediate budgets. These results indicate that single-turn
performance is insufficient to characterize interactive coding ability.
\end{abstract}

\section{Introduction}
\label{sec:intro}

\begin{figure}[!ht]
    \centering
    \includegraphics[width=\columnwidth]{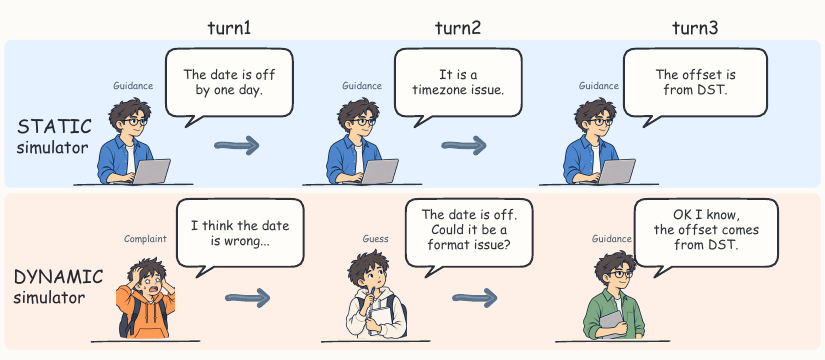}
    \caption{Existing multi-turn benchmarks keep the simulated user in one persona, while in the real development sessions we measure a developer's style shifts from turn to turn.
    On the same date-off-by-one bug, the static simulator phrases every turn as precise guidance, while the \dataname simulator opens on a vague complaint, guesses a wrong cause, and reaches guidance only at the third turn.}
    \label{fig:dynamic-user-teaser}
\end{figure}

AI programming assistants are deployed at scale and used daily by developers \citep{liang2024survey}.
Their form has evolved from autocompletion to context-grounded conversation, and then to agents that navigate repositories, edit files, and run tests on their own — a practice popularly called vibe coding \citep{ross2023assistant, sweagent, vibecoding}.
Developers pose follow-up questions and refine their requests in response to intermediate model outputs \citep{barke2023grounded,ross2023assistant}, where initial requests are often incomplete and gradually clarified across turns. Studies of agents in real-world settings further confirm this same iterative pattern holds across entire tasks \citep{vibecodingmlr,swechat}.
As a representative scenario, debugging with an agent has now become a standard conversational workflow for developers.

However, prevailing benchmarks do not measure how well an agent works in this scenario for two main reasons.
One is that most use a single-turn setup, handing the agent everything at once and asking for a correct patch in return, so the agent is never put in the conversation where the work actually happens.
The few that do run several turns stay static, releasing the information in a written order or phrasing every turn in one fixed conversational style, which leaves the agent talking to an unresponsive mechanism and not to a simulation of a developer.
Figure~\ref{fig:dynamic-user-teaser} sets a static user beside a dynamic one on the same task.
The other concerns multi-turn judging, where every intermediate turn calls for independent assessment, but standard tests can only provide factual execution evidence, and unrestricted LLM judges are prone to making inferences beyond what results actually demonstrate.
For both reasons, a benchmark that measures coding agents' abilities in conversation is still missing.

To address this, we propose \dataname, which settles where the tasks come from, where the turns come from, and how each turn is judged.
The first is a \textbf{construction pipeline} that produces multi-turn tasks along two routes.
One route mines a repository's own history for paired issues that a maintainer publicly recorded as linked.
The other takes an instance of an existing single-turn benchmark and stages it as a conversation.
As we continue running the pipeline along these two routes, it consistently generates new and up-to-date tasks.
The second is a method for \textbf{simulating the user} across the turns of a conversation.
We calculate a persona transition matrix on \nsessions real development sessions from an internally deployed commercial coding assistant, and we write each turn's utterance from that matrix together with the information pools of the task, so the persona an agent meets becomes dynamic as the session goes on.
For each task we draw the whole chain in advance, which ensures any agent on the same task meets the same user.
The third is an \textbf{evaluation framework}, which enables the coding agent to operate under constraints, with dual agents judging its generated outputs based on evidence.
The framework sets rules that prevent coding agents from receiving credit beyond their demonstrated ability.
Based on tests run after every turn, a Judge from one family writes the verdict, and a Reviewer from another family clears it before it counts.
We select 72 tasks from the pipeline as \dataname and evaluate agents on it.

The evaluation on \dataname reveals behavior that a single-turn evaluation cannot show.
Claude Fable 5.1 resolves the most tasks within a single turn, 31.9\%, and still finishes second on the length-discounted pass rate, behind Claude Opus 5, which reaches a similar place in fewer turns, 2.99 against 3.14.
GPT-6-Astra resolves the most tasks of any agent by the eighth turn, 81.9\%, and places fourth once the turns it spends getting there are discounted.

In this paper, we make the following contributions.
\textbf{(1) A construction pipeline.}
It builds multi-turn tasks from repository evidence and existing single-turn benchmarks.
\textbf{(2) A multi-turn user simulation.}
We introduce a user simulator with dynamic personas grounded in real development conversations.
\textbf{(3) An evaluation framework.}
We propose a dual-agent judging evaluation framework complemented by a suite of supporting mechanisms, such as equivalent pass testing.
\dataname will be released along with the evaluation framework.

\section{Related work}
\label{sec:related}
Table~\ref{tab:compare} compares \dataname with existing coding-agent benchmarks. Below, we discuss two complementary lines of related work.

\paragraph{Multi-turn benchmarks and simulated users.}
Multi-turn evaluations must specify both what the user reveals and how the user speaks. SWE-Interact releases a task's requirements one at a time and uses a fixed, terse senior-developer persona derived from SWE-chat \citep{sweinteract,swechat}, while SWE-Together reconstructs a task-specific simulator from a recorded session and triggers utterances according to the live trajectory \citep{swetogether}. These approaches condition information release on the trajectory but do not vary the user's voice according to a measured distribution of developer personas. More broadly, simulated users can diverge from real developers, and persona choice can substantially change measured performance \citep{lostinsimulation,sim2real,personaaug}. \dataname samples a persona chain from real development sessions and generates an information pool for each task, which helps simulate users naturally.

\paragraph{SWE benchmarks and evaluation reliability.}
SWE-bench established issue-level, execution-based evaluation for coding agents \citep{swebench}. Later work addresses task contamination and freshness through larger, private, or continuously refreshed task collections \citep{swebenchplus,swebenchillusion,swebenchpro,swebenchlive}; SWE Context Bench is closest to our mining route but poses related instances independently \citep{swecontextbench}. A parallel line shows that verdicts can fail because tests are too loose or tight, LLM judges exhibit bias, or the evaluation harness can be gamed \citep{swebenchplus,zheng2023judging,wataoka2024selfpreference,benchjack}. Panel-based judges and rule-based checkers improve reliability but remain imperfect \citep{poll,agentrewardbench}. In the multi-turn setting, SWE-Interact still verifies only the final solution and rubric-scores intermediate turns \citep{sweinteract}. \dataname executes tests after every turn and uses a cross-family writer/reviewer gate, grounding each verdict in execution and reserving model judgment for evidence the tests cannot cover.

\begin{table}[ht!]
\centering
\caption{A comparison of recent benchmarks \citep{swebench, swebenchpro,
swebenchlive, sweevo, swecontextbench, sweinteract, swetogether} with
\dataname.}
\label{tab:compare}
\footnotesize
\setlength{\tabcolsep}{2.7pt}
\begin{tabular}{@{}lccc>{\raggedright\arraybackslash}p{4.2cm}c@{}}
\toprule
\textbf{Benchmark} & \textbf{Recent tasks} & \textbf{Multi-turn} & \textbf{Persona}
  & \textbf{Judge method} & \textbf{Judge freedom} \\
\midrule
SWE-bench
  & $\times$ & $\times$ & $\times$ & tests & Low \\
SWE-bench Pro
  & $\times$ & $\times$ & $\times$ & tests, hand-checked & Low \\
SWE-bench-Live
  & \checkmark & $\times$ & $\times$ & tests, auto-validated & Low \\
SWE-Evo
  & $\times$ & $\times$ & $\times$ & tests & Low \\
SWE-ContextBench
  & $\times$ & $\times$ & $\times$ & tests & Low \\
SWE-Interact
  & $\times$ & \checkmark & static & source verifier & Low \\
SWE-Together
  & \checkmark & \checkmark & static & rubric judge & High \\
\midrule
\textbf{\dataname}
  & \checkmark & \checkmark & dynamic & two judges on test evidence
  & Medium \\
\bottomrule
\end{tabular}

\vspace{1.5mm}
\parbox{\textwidth}{\footnotesize\itshape Persona records whether the
conversational persona stays fixed across turns.\\ Judge freedom records how far
the verdict rests on an LLM: Low means no LLM takes part in deciding it, High
means it is decided entirely by an LLM, and Medium means the two are combined.}
\end{table}

\section{Preliminaries}
\label{sec:setting}

The dominant executable format for software-engineering evaluation poses one issue as one task and grades the returned patch by executing tests \citep{swebench,swebenchpro}.
Given an issue and the repository state $b$ at the commit preceding its fix, the agent answers once with a patch $\hat g$.
Write $t(s)\in\{0,1\}$ for the outcome of test $t$ in workspace state $s$, and let $s\oplus p$ apply patch $p$ to $s$.
The task also carries the maintainer's own fix, the \emph{gold patch} $g$, and two sets of tests from $b$ and $b\oplus g$,
\begin{equation}
t(b)=0 \ \ \text{and} \ \ t(b\oplus g)=1 \quad \forall\, t\in F,
\qquad\qquad
t(b)=t(b\oplus g)=1 \quad \forall\, t\in P,
\label{eq:transition}
\end{equation}
the \emph{fail-to-pass} set $F$, which the fix has to turn green, and the \emph{pass-to-pass} set $P$, which it must not break.
A patch $\hat g$ \emph{resolves} the task when $t(b\oplus\hat g)=1$ for every $t\in F\cup P$, and the reported score is the fraction of tasks resolved \citep{sweagent}.

\section{\dataname}
\label{sec:bench}
\begin{figure}[t]
\centering
\includegraphics[width=\linewidth]{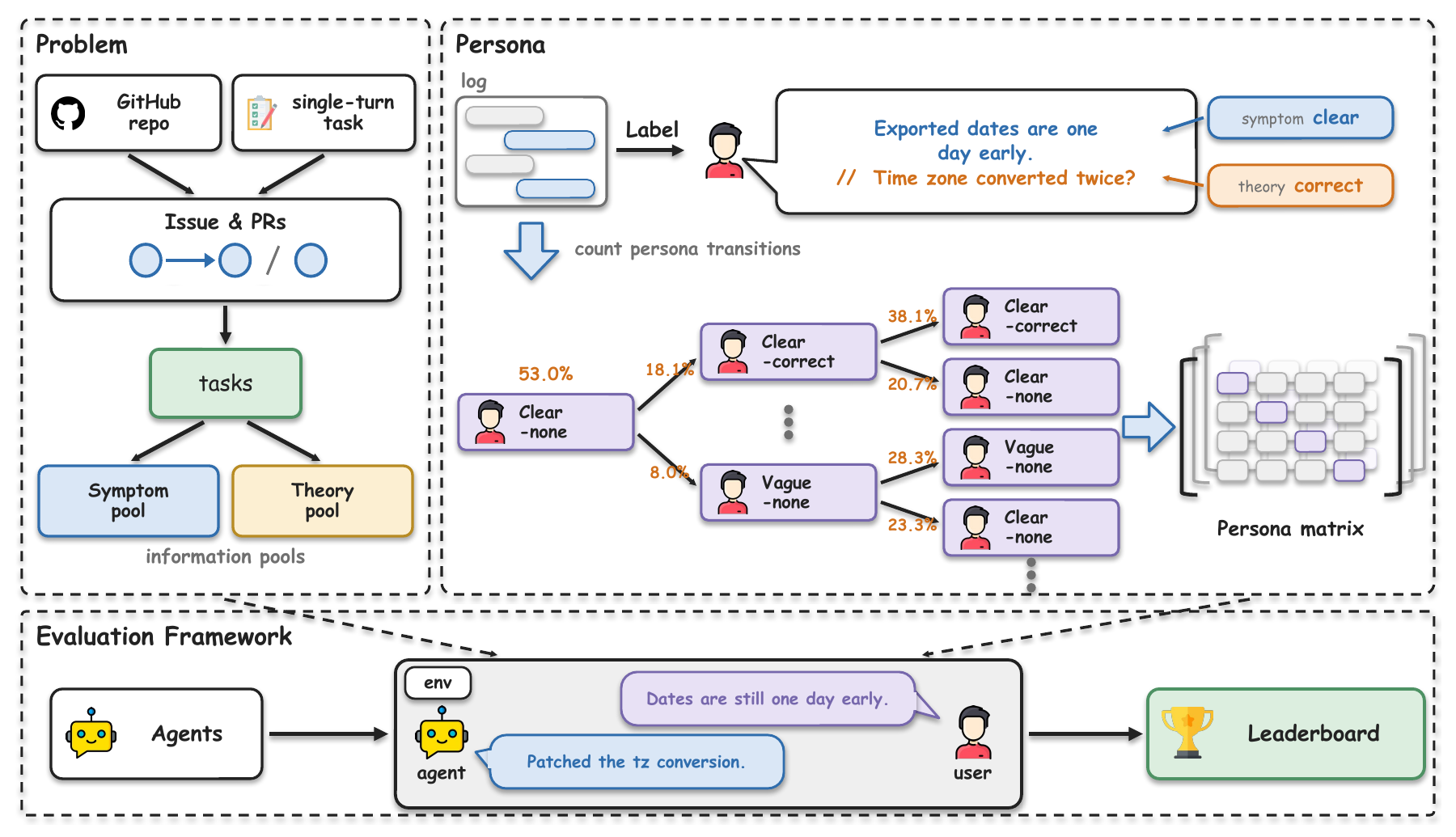}
\caption{The \dataname pipeline. The problem panel shows how a task is
constructed, the persona panel how the simulated user is built, and the
evaluation panel how a session is run and scored.}
\label{fig:overview}
\end{figure}

We introduce \dataname, which evaluates agents on issue fixing inside a multi-turn conversation and so measures coding ability more completely than a one-shot format can.
Where Section~\ref{sec:setting} is settled in a single exchange, an agent here works on the issue over several turns, each turn answering what the user has just asked for, and we check the fail-to-pass and the pass-to-pass set after every one of those turns.
Turning both sets green in as few turns as possible is what \dataname credits as strength at resolving code issues interactively.
Figure~\ref{fig:overview} lays out in order where the tasks come from, how the multi-turn form is designed, and how each turn is judged.

\subsection{Data Construction}
\label{sec:data}
We mine issue cascades and single issues from two sources and select the suitable ones as tasks.
The first route searches repository histories for two issues explicitly linked by a maintainer.
Each side has a merged fix with tests, giving the task two related acceptance points.
The second route takes tasks from published single-turn benchmarks, preserving their repository state, gold patch, tests and original acceptance point while changing how their information is released across a multi-turn conversation.

Both routes then turn each task into an information pool.
The pipeline extracts symptoms from the issue report and uses the gold patch to construct correct or wrong directions about the cause, with the wrong direction later shared by the off-target and wrong personas.
A cascade also carries follow-up information for its second issue.
The pool is frozen before evaluation and checked for identifiers introduced by the gold patch, with an additional cross-phase scope check for cascades.
Agents evaluated on the same task therefore draw from the same frozen candidate information pool, preventing differences in pool construction from confounding the comparison.

Before admission, each task passes both mechanical and human checks.
The framework reruns the fail-to-pass and pass-to-pass tests in the evaluation container, verifying that the fail-to-pass tests fail at the base state and pass after the gold patch is applied, while the pass-to-pass tests pass in both states.
Appendix~\ref{app:construction} details how tasks from both sources are converted into information pools and admitted as executable tasks.
\subsection{Multi-turn Interaction}
\label{sec:multiturn}
Section~\ref{sec:data} fixes what the simulated user may say, but not which item should be selected at each turn or the voice in which it should be expressed.
We handle these questions separately, first selecting the content through a release policy and then rendering it through a conversational persona model calibrated on real development sessions.

After each turn, the simulated user selects guidance from the information pool according to the agent's response and the canonical test results.
The information selection process follows a fixed priority hierarchy designed to mirror the logic of natural developer interactions. If the agent poses a direct question, the simulated user responds with a direct answer. Otherwise, it selects content in descending order of priority: the most severe user-observable blocking failure, an unaddressed core business boundary, or a secondary edge case. This hierarchy governs what information the user simulator releases at the current turn.

Once the content is selected, the framework must decide how to express it.
The expression of the developer varies between people and between turns of the same session.
Its two common components align with the two kinds of content in the information pool, namely how a developer describes \emph{what was observed} and whether they offer \emph{a view of the cause}.
We therefore factor the simulated user along two axes.
Clarity records whether the symptom is stated precisely or vaguely, and veracity records whether the user offers no view, a correct one, an off-target one, or a wrong one.
Crossing these axes yields eight conversational personas, and Table~\ref{tab:voice} presents the same request phrased in each of them.

\begin{table}[t]
\centering
\caption{How one symptom, dates coming out of the export a day early because
the timezone conversion runs twice, is put under the eight personas. Columns
give the four ways of stating a cause, rows the two ways of describing the
symptom.}
\label{tab:voice}
\footnotesize
\begin{tabular}{@{}l *{4}{>{\raggedright\arraybackslash}p{0.19\linewidth}}@{}}
\toprule
 & \textbf{no cause} & \textbf{correct} & \textbf{off-target} & \textbf{wrong} \\
\midrule
\textbf{clear}
  & Exported dates are one day early.
  & Exported dates are one day early. \slash\slash\ Timezone converted twice?
  & Exported dates are one day early. \slash\slash\ Maybe the template?
  & Exported dates are one day early. \slash\slash\ Timezone is fine, it is
    the template. \\
\addlinespace
\textbf{vague}
  & The export looks off.
  & The export looks off. \slash\slash\ Timezone converted twice?
  & The export looks off. \slash\slash\ Maybe the template?
  & The export looks off. \slash\slash\ Timezone is fine, it is the
    template. \\
\bottomrule
\end{tabular}
\end{table}

We do not assume the eight cells occur uniformly.
From \nsessions real development sessions from an internally deployed commercial coding assistant, we measure the opening persona and its turn-to-turn transitions.
Of these sessions, 13,530 change code, and the eight cells account for 98.65\% of them, with 52.98\% opening on a clear symptom and no view of the cause. 
The transitions form a family of matrices conditioned on the opening persona rather than one global matrix.
We pseudorandomly draw a complete persona chain for each task in advance so that different agents on the same task face the same changes in the user.
The selected content is rendered in the persona assigned to that turn.
Appendix~\ref{app:personas} gives all eight opening-conditioned transition matrices and the persona design.

The information pool fixes what may be said, the release policy selects what this turn says and the frozen persona chain fixes how it is said, which completes the multi-turn interaction.
What remains is how each turn is judged, which is the next subsection.

\subsection{Evaluation}
\label{sec:eval}

A multi-turn evaluation framework entails two dimensions of freedom, each causing evaluation misjudgments when improperly calibrated.
The degree of freedom allowed by the framework, when set too loose or too tight, renders the evaluation scores unreliable.
The degree of freedom granted to the agent, when set too loose or too tight, prevents it from demonstrating its true capabilities.
Our evaluation framework for \dataname addresses these two dimensions separately.

\begin{algorithm}[t]
\caption{Scoring one session.}
\label{alg:eval}
\footnotesize
\begin{algorithmic}[1]
\Require coding agent under test $M$, task $\tau$, persona seed $\sigma$, turn
budget $K$
\Ensure $\mathrm{LDPR}\in[0,1]$
\State $\textsc{Judge} \leftrightarrow \textsc{Review}$
       \Comment{two different model families, swapped between tasks}
\State $\mathit{workspace} \gets$ base state of $\tau$
\State $\mathit{message} \gets$ opening request of $\tau$ under $\sigma$
\State $\mathit{result} \gets$ every acceptance point of $\tau$, pass-to-pass
       green and fail-to-pass red
\State $k \gets 0$
\While{$k < K$ \textbf{ and } $\mathit{result}$ is not all-pass}
  \State $\mathit{previous} \gets \mathit{workspace}$
  \State $\mathit{workspace} \gets M(\mathit{message}, \mathit{workspace})$
  \State $k \gets k + 1$ if $\mathit{workspace}\ne\mathit{previous}$, else
         $k \gets k + 0.5$
  \State $\mathit{result} \gets \textsc{RunTest}(\mathit{workspace},
         \mathit{result})$
         \Comment{execute the tests in the container}
  \State $(\mathit{scorecard}, \mathit{message}) \gets
         \textsc{Judge}(\mathit{workspace}, \mathit{result}, \sigma,
         k)$
         \Comment{verdict, and the next request}
  \State $\mathit{result} \gets \textsc{EqualPass}(\mathit{workspace},
         \mathit{result}, \mathit{scorecard})$
         \Comment{run the rewritten test where one is asked for}
  \State $\mathit{label} \gets \textsc{Review}(\mathit{scorecard},
         \mathit{message}, \mathit{result})$
  \If{$\mathit{label} = \textit{rewrite}$}
    \State $(\mathit{scorecard}, \mathit{message}) \gets$ the rewrite
           \textsc{Review} returns
  \EndIf
\EndWhile
\State \Return $(K + 1 - k)/K$ if $\mathit{result}$ is all-pass,
       else $0$
\end{algorithmic}
\end{algorithm}

\paragraph{Freedom for the agent.}

Deciding whether a solution is correct can fail in either direction.
Tests are written around the maintainer's own fix, so they may reject a patch that implements the behavior without converging on the interface of the gold patch, producing a false negative.
In a neighboring setting this is measurable, where rule-based evaluation of agent trajectories recovers only 55.9\% of the runs an expert calls successful, because the check demands an exact match and turns away trajectories that differ in ways that do not matter \citep{agentrewardbench}.
At the other end, tests weak enough to pass a patch that fixed nothing give a false positive \citep{swebenchplus}.
Moving the primary signal off test execution and onto a model's reading of whether the semantics are satisfied \citep{swetogether} widens the false positives.
Admitting only the verbatim passing of the shipped tests \citep{swebench, swebenchpro} widens the false negatives.
In \dataname the framework may, on the basis of what the agent actually implemented, propose a change to a test and count a pass on the genuinely modified test as an \textbf{equivalent pass}, which is \textsc{EqualPass} in Algorithm~\ref{alg:eval}.
How much capability it brings to bear is likewise not capped, since a setting that gives no tools and asks for a patch from the issue body alone cannot measure the difference between scaffolds \citep{swebench, sweagent}.
We block its access to the assets that would settle the task for it, and restrict outside lookup where that is necessary.

\paragraph{Freedom for the instrument.}

The freedom given to the simulated user and the Judge does not decide what is correct.
It decides whether the instrument has itself become a variable.
A simulated user left to improvise drifts with whichever model it runs on, so one task stops being the same task across two evaluations \citep{sweinteract}.
A simulated user whose every line is written in advance is not a user.
We freeze the agenda and not the wording.
Which persona holds at which turn and which facts may be said are fixed by the persona chain and the information pools, and the opening request is fixed outright so that every agent starts from the same point.
Which fact to raise at a later turn and how to phrase it is left to the simulated user.
The Judge side is the same problem seen from the other end.
A language model scores its own family's output too highly \citep{zheng2023judging, wataoka2024selfpreference}, and a panel drawn from disjoint families brings that bias down \citep{poll}, so we chain two families.
The Judge writes a verdict and a Reviewer from a different family may rewrite it \emph{before} it takes effect, with the two swapped between tasks.
This is an inline gate rather than a post-hoc audit, and every judgment the framework issues has passed through both.

Both halves rest on the same thing.
A verdict counts only where the tests have been run and have said so, since the outcome the loop returns is the outcome of executing them, and a model's reading enters where the tests are silent rather than where they disagree with it.

\paragraph{Metrics.}

A multi-turn setting admits a quantity the one-shot format has no room for, namely how much of the turn budget an agent had left when it succeeded.
Beyond the plain resolve rate we therefore report the \emph{L}ength-\emph{D}iscounted \emph{P}ass \emph{R}ate (LDPR), which is counted as
\begin{equation}
\mathrm{LDPR} \;=\;
\begin{cases}
(K + 1 - k)/K & \text{tests all pass,}\\[2pt]
0             & \text{otherwise,}
\end{cases}
\label{eq:ldpr}
\end{equation}
where $k$ is the weighted number of turns used; a turn without a code edit counts as half a turn.
Agents that resolve a task in fewer conversational turns receive higher scores, while those that fail to complete the task within the predefined maximum turn limit are assigned a score of zero.

Algorithm~\ref{alg:eval} shows how an agent exhibits its ability within the evaluation framework, in interaction with \textsc{Judge} and \textsc{Review}.
The more detailed algorithmic logic and the framework's actual implementation are in Appendix~\ref{app:scoring}.

\section{Experiments}
\label{sec:exp}

\begin{table}[t]
\centering
\caption{Evaluation results on \dataname. The middle columns give the resolved
rate within that many turns.}
\label{tab:leaderboard}
\small
\setlength{\tabcolsep}{4pt}
\setlength{\aboverulesep}{0pt}
\setlength{\belowrulesep}{0pt}
\setlength{\extrarowheight}{.6ex}
\begin{tabular}{@{}l|cccccccc|cc@{}}
\toprule
& \multicolumn{8}{c|}{Resolved within $k$ turns} & & \\
\cmidrule(l{-0.5pt}){2-9}
Model & 1 & 2 & 3 & 4 & 5 & 6 & 7 & 8 & $k$ & LDPR \\
\midrule
Claude Opus 5 & 25.0 & \textbf{56.9} & \textbf{66.7} & \textbf{76.4} & \textbf{79.2} & \textbf{80.6} & \textbf{80.6} & 80.6 & \textbf{2.99} & \textbf{0.640} \\
Claude Fable 5.1$^{*}$ & \textbf{31.9} & 47.2 & 62.5 & 66.7 & 75.0 & 75.0 & 75.0 & 76.4 & 3.14 & 0.593 \\
Grok-4.6 & 18.1 & 43.1 & 58.3 & 66.7 & 76.4 & 79.2 & 79.2 & 80.6 & 3.40 & 0.569 \\
GPT-6-Astra & 12.5 & 38.9 & 55.6 & 70.8 & 77.8 & 79.2 & \textbf{80.6} & \textbf{81.9} & 3.44 & 0.561 \\
DeepSeek-v4.1-Flash & 18.1 & 40.3 & 51.4 & 65.3 & 73.6 & 76.4 & 76.4 & 76.4 & 3.47 & 0.537 \\
GPT-5.6-Sol & 16.7 & 36.1 & 47.2 & 61.1 & 68.1 & 69.4 & 70.8 & 70.8 & 3.73 & 0.500 \\
GLM-5.3 & 19.4 & 37.5 & 51.4 & 63.9 & 63.9 & 65.3 & 65.3 & 65.3 & 3.74 & 0.496 \\
Gemini-3.8-Flash & 15.3 & 33.3 & 44.4 & 54.2 & 62.5 & 62.5 & 63.9 & 66.7 & 3.79 & 0.458 \\
Hunyuan-4.0-preview & 9.7 & 30.6 & 48.6 & 56.9 & 62.5 & 63.9 & 65.3 & 66.7 & 3.85 & 0.457 \\
Doubao-Seed-2.1-Pro & 11.1 & 31.9 & 43.1 & 54.2 & 68.1 & 68.1 & 68.1 & 68.1 & 3.89 & 0.456 \\
Kimi-K3 & 18.1 & 27.8 & 36.1 & 44.4 & 51.4 & 54.2 & 55.6 & 56.9 & 4.14 & 0.391 \\
\bottomrule
\end{tabular}
\vspace{1.5mm}
\parbox{\linewidth}{\footnotesize\itshape\raggedright $^{*}$Claude Fable 5.1
refused some tasks. We count a refusal as unresolved, so its reported rates carry that loss.}
\end{table}

We evaluate how current agents fix issues in multi-turn interaction on \dataname, and then we analyze the multi-turn design and the evaluation framework.

\subsection{Main results}
\label{sec:mainresults}
We select 72 tasks from the continuously running pipeline.
Appendix~\ref{app:construction} details task construction and admission.
We evaluate 11 agents on \dataname, each a model on a scaffold as listed in Table~\ref{tab:systems-scaffolds}. Of these tasks, 22 are mined from open-source repository history and contain two issues, while the other 50 are converted from published single-turn benchmarks.

The evaluation framework uses two different model families for user simulation and result adjudication.
The judging and review sides are rotated between gpt-5.6-sol on the Codex scaffold and opus-5 on the Claude Code scaffold.
We selected them because they represent the most capable cross-family model combination that can be deployed stably at present. The framework is not bound to these two specific models, and either can be replaced with suitable alternatives.
Information pools and persona trajectories are drawn per task in advance, so every agent is evaluated fairly on the same task.

The two subsets run under different turn budgets, $K=5$ for a single-issue task and $K=8$ for a two-issue one, since a second acceptance point needs the extra turns to be reached at all.
Table~\ref{tab:leaderboard} reports results on \dataname, the share of tasks resolved within the first $k$ turns for $k$ from 1 to 8, together with LDPR (Equation~\ref{eq:ldpr}) and the mean number of turns used.
More detailed results, such as costs, are in Appendix~\ref{app:percell}, and Appendix~\ref{app:cases} follows individual sessions case by case.
Appendix~\ref{app:stats} reports confidence intervals and paired comparisons between agents.

\subsection{Analysis of the multi-turn design}
\label{sec:val:multiturn}

\subsubsection{Multi-turn against single-turn}
\label{sec:val:oneshot}

As the main table shows, the agent that is best at the opening is not the agent that is best at the end.
Claude Fable 5.1 resolves more tasks within the first turn than any other agent and still finishes behind Claude Opus 5 on LDPR.
GPT-6-Astra resolves more tasks by the eighth turn than any other agent and places fourth on LDPR.
A single-turn evaluation cannot show either of these, because it observes one point of the curve and reports it as the whole of an agent's ability.

\begin{figure}[h]
\centering
\includegraphics[width=\linewidth,trim=0 18pt 0 0,clip]{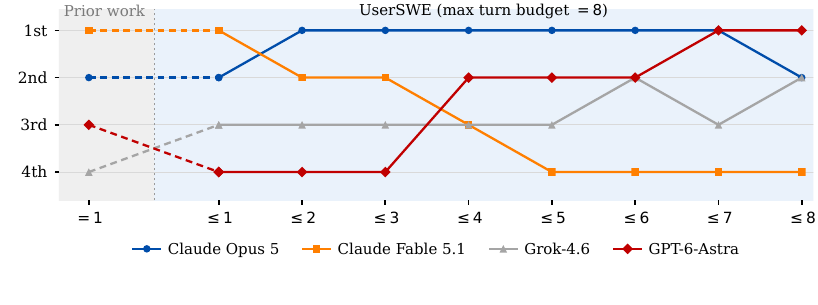}
\caption{The order of the four leading agents on published single-turn tasks, and how that order changes as the turn budget grows on \dataname.}
\label{fig:turnrank}
\end{figure}

We look more closely at the four agents that lead on LDPR.
Figure~\ref{fig:turnrank} tracks the order they stand in as the turn budget grows.
Over eight budgets the four take six different orderings between them, and three of the four lead at some budget.
GPT-6-Astra is last of the four through the third turn and catches up to lead by the eighth.
They end within six points of each other, between 76.4\% and 81.9\% resolved by the eighth turn, so these are reorderings among agents of comparable final ability.
We also compare these four agents on a separate collection of 42 single-turn tasks drawn from SWE-bench Pro and SWE-bench-Live.
Claude Fable 5.1 ranks first at 78.6\%, well ahead of Claude Opus 5 at 64.3\%, GPT-6-Astra at 61.9\% and Grok-4.6 at 59.5\%.
Fable's large single-turn margin explains its early lead, but the ordering changes as the turn budget grows.

\subsubsection{Dynamic conversational persona}
\label{sec:val:persona}

Most existing benchmarks model the simulated user as a static persona with both clear expression and correct causal views \citep{sweinteract,swechat}. 
That profile opens only 19.03\% of the 13,530 observed sessions, and holding it for a whole session is rarer still: 0.72\% of sessions stay clear and correct for five consecutive turns on the same issue and 0.24\% do so for eight. 
Even among the sessions that actually run that long, the shares are 2.81\% and 1.86\%.

We compare a reference arm that remains clear and correct at every turn with the persona chain. Both arms run the 50 single-issue tasks, which fit a five-turn budget and so isolate the persona from the cascade's issue-to-issue carryover, and each comparison uses the same task and agent and changes only the user trajectory. Table~\ref{tab:personaablate} reports LDPR and mean weighted turns $k$.
Three of the four agents move the same way, scoring higher and finishing sooner under the clear-and-correct user, and the size of the shift varies widely between them, from 0.020 to 0.076 in LDPR.
The largest is GPT-6-Astra, whose LDPR increases from 0.580 to 0.656 while $k$ decreases from 2.92 to 2.54, and Kimi-K3 moves the other way.
How much an agent depends on precise phrasing is itself a property of the agent, and it shows across all 72 tasks and all eleven agents. Splitting the full set by whether the opening states the symptom precisely or as an impression, the gap between the two halves runs from 0.268 for Claude Fable 5.1 down to $-$0.088 for DeepSeek-v4.1-Flash (Appendix~\ref{app:clarity}).
An idealized user therefore reports a benchmark easier than the one most users present, and it flatters some agents more than others.
\begin{table}[t]
\centering
\caption{Persona ablation under a clear-and-correct user and the dynamic
persona chain, over the same 50 single-issue tasks. $\Delta$ is the change from
clear-and-correct to dynamic.}
\label{tab:personaablate}
\small
\setlength{\tabcolsep}{4pt}
\begin{tabular*}{\textwidth}{@{\extracolsep{\fill}}lcccccc@{}}
\toprule
 & \multicolumn{3}{c}{LDPR} & \multicolumn{3}{c}{$k$} \\
\cmidrule(lr){2-4}\cmidrule(lr){5-7}
Model & Clear & Dynamic & $\Delta$ & Clear & Dynamic & $\Delta$ \\
\midrule
Claude Opus 5       & 0.688 & 0.668 & $\downarrow$.020 & 2.27 & 2.50 & $\uparrow$.23 \\
GPT-6-Astra         & 0.656 & 0.580 & $\downarrow$.076 & 2.54 & 2.92 & $\uparrow$.38 \\
DeepSeek-v4.1-Flash & 0.628 & 0.562 & $\downarrow$.066 & 2.58 & 2.88 & $\uparrow$.30 \\
Kimi-K3             & 0.428 & 0.470 & $\uparrow$.042 & 3.46 & 3.25 & $\downarrow$.21 \\
\bottomrule
\end{tabular*}
\end{table}

\subsection{Analysis of the evaluation framework}
\label{sec:val:framework}

The rates and cases in this section come from all canonical runs of the eleven
leaderboard models, across the 22 cascade tasks and 50 single-issue tasks.
These runs used the same Judge, Reviewer and equivalence protocol, so they can
be read for what the framework itself did.
We report them only as proportions.

\subsubsection{Reviewer rewrites}
\label{sec:val:reviewer}

The Reviewer is not a formality.
Across the scored turns, it rewrote 40.9\% of Judge decisions.
Table~\ref{tab:reviewercat} separates the two Reviewer configurations.
It reports the share of reviewed turns that each Reviewer rewrote and how the objections attached to those rewrites divide across four review domains.
\begin{table}[h]
\centering
\caption{Reviewer rewrites and their objections.  Rewrite is the share of
reviewed turns rewritten within each configuration; the four objection columns
break down that configuration's own objections and sum to 100\% across each
row.}
\label{tab:reviewercat}
\small
\setlength{\tabcolsep}{5pt}
\renewcommand{\arraystretch}{1.08}
\begin{tabular*}{\linewidth}{@{\extracolsep{\fill}}l!{\color{black!35}\vrule width 0.35pt}c!{\color{black!35}\vrule width 0.35pt}cccc@{}}
\toprule
Reviewer & Rewrite & Evidence & Guidance & Equivalence & Persona \\
\midrule
Codex / GPT-5.6-Sol & 62.4\% & 70.3\% & 14.3\% & 9.5\% & 5.9\% \\
Claude Code / Claude Opus 5 & 25.4\% & 42.1\% & 41.5\% & 5.3\% & 11.1\% \\
\bottomrule
\end{tabular*}
\end{table}

The first column is calculated over reviewed turns within each configuration; the other four partition that configuration's own objections, so each row of the four sums to 100\%.
Evidence marks a state unsupported by per-item test results or an executed command, guidance a next-turn utterance that breaks the interaction rules, equivalence a mishandled equivalence request, and persona a turn that does not execute the trajectory drawn for it.
Both configurations catch problems that would otherwise take effect under a single Judge.
The two configurations object to different things: evidence accounts for 70.3\% of what the Codex Reviewer files, whereas the Claude Code Reviewer splits its objections almost evenly between evidence and guidance, 42.1\% and 41.5\%.
Overall, 87.4\% are rated major or critical, meaning that under the rubric they may change the follow-up, completion state or score rather than wording alone.
Appendix~\ref{app:reviewer-models} gives the rewrite rate and objection distribution for every leaderboard model.

\subsubsection{Equivalent passes}
\label{sec:val:equiv}

Our equivalent-pass mechanism is strictly execution-based. Tests tied to the gold patch's interface or structure may reject functionally equivalent implementations. We allow targeted rewrites for the tested implementation, but award credit only when the rewritten test passes and still fails at baseline; descriptive claims do not count.

\begin{figure}[h]
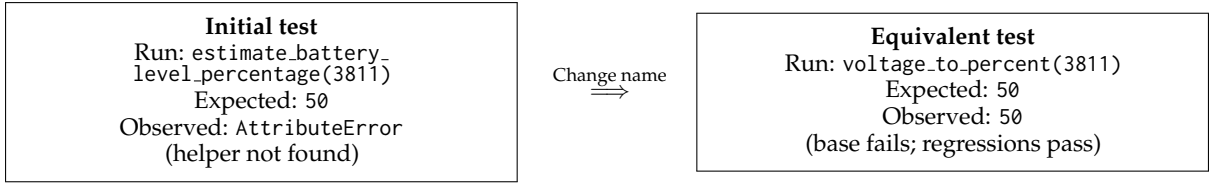

\centering
\small
\setlength{\fboxsep}{5pt}
\fbox{\parbox[c]{0.40\linewidth}{\centering
\textbf{Initial test}\\
Run: \texttt{estimate\_battery\_}\\[-2pt]
\texttt{level\_percentage(3811)}\\
Expected: \texttt{50}\\
Observed: \texttt{AttributeError}\\
(helper not found)}}
\hfill$\overset{\text{\scriptsize Change name}}{\Longrightarrow}$\hfill
\fbox{\parbox[c]{0.40\linewidth}{\centering
\textbf{Equivalent test}\\
Run: \texttt{voltage\_to\_percent(3811)}\\
Expected: \texttt{50}\\
Observed: \texttt{50}\\
(base fails; regressions pass)}}
\caption{The equivalent test retains the expected behavior under a different helper name.}
\label{fig:equivpass}
\end{figure}

The route is used and its approval rate does not make review a formality.
Across these runs 14.6\% of cells filed at least one equivalence request, and 51.2\% of the requests they filed were granted and written into the scorecard.
Figure~\ref{fig:equivpass} gives one such grant without reproducing code.
The shipped test called the battery estimator by the helper name used in the gold patch, while the candidate implemented the same 22 voltage-to-percentage mappings under a different name.
The rewrite retained every input-output assertion but located the one-argument estimator by its signature; it failed at the base state, passed on the candidate without regression, and was approved by the Reviewer.
Where the other requests were stopped is broken down in Appendix~\ref{app:artifacts}.

\section{Limitations}
\label{sec:limitations}

\dataname has three limitations.
\textbf{(1) Scale.} Multi-turn evaluation runs a full session and reruns tests after every turn, making it more expensive than scoring one patch per task and limiting the released set to 72 tasks. Table~\ref{tab:percell_cost} reports detailed per-task costs.
\textbf{(2) Horizon.} \dataname covers one maintenance episode over one or two issues, from a base commit to a fixed set of acceptance points. It does not cover longer-term work such as building a repository from scratch, which remains a direction for future extensions.
\textbf{(3) Session logs.} The user model uses opening distributions and turn-to-turn transitions measured from \nsessions sessions collected from an internally deployed commercial coding assistant. Future versions could incorporate additional signals about task types and request phrasing.

\section{Conclusion}
\label{sec:conclusion}
We present \dataname, a benchmark for evaluating coding agents in multi-turn
conversations with dynamic users. We contribute a construction pipeline that builds
multi-turn tasks from repository history and existing benchmarks, a user
simulator with dynamic personas grounded in real development sessions, and an
evaluation framework that anchors cross-family judging and review to executed
tests. Together, these components produce \dataname, which we use to evaluate
eleven agents on 72 tasks and uncover behaviors that single-turn
evaluation cannot reveal.

\subsection*{AI use statement}

We used generative AI tools to polish the manuscript and as necessary
components of the experimental framework, as described in the paper.
The authors reviewed all AI-assisted work and take full responsibility
for the final content and reported results.

\subsection*{Ethics statement}

The benchmark tasks are constructed from publicly available issues, pull requests and repository histories, together with previously published benchmarks. Repository-derived materials remain subject to their upstream licenses. 
The user model is calibrated from recorded development sessions collected from an internally deployed commercial coding assistant. We do not release raw conversations or person-level attributes, and release only the aggregate statistics required to instantiate the simulator. 
This limits the released behavioral information to what is needed to instantiate the simulator while reducing privacy and security risks associated with development logs.

\subsection*{Reproducibility statement}

We document the benchmark construction pipeline
in Section~\ref{sec:data} and Appendix~\ref{app:construction}.
Appendix~\ref{app:personas} specifies the simulated user, and
Appendix~\ref{app:transition} reports the calibrated opening distribution and
transition matrices. Judge states, Reviewer actions and equivalence settlement
are specified in Appendix~\ref{app:scoring}, and the post-run audit rubric is
provided in Appendix~\ref{app:rubric}. Agent configurations and cost
measurements appear in Appendix~\ref{app:percell}. The sampled persona trajectories are
independent of the agents and are stored to permit turn-by-turn
replay. We will release \dataname together with the multi-turn evaluation
framework.

\bibliographystyle{iclr2027_conference}
\bibliography{refs}

\appendix
\section{How a task is built}
\label{app:construction}

The released benchmark contains 22 cascades mined from repository history and
50 single-issue tasks converted from published benchmarks. This appendix
follows both routes from source material to a task package.

\subsection{Two source routes}

The cascade route searches repository text for two signals.
Reveal signals occur when a later issue says it was exposed while investigating
or fixing an earlier issue.
Dependency signals occur in pull-request text that states one change is stacked on, depends on or follows another.
A pair is retained in the stated order only when both issues have merged fixes
and the later record points to the earlier work. Of the 22 mined cascades, six
come from reveal signals and sixteen from dependency signals.

The separability check in Section~\ref{app:admission}, rather than the phrase
category, establishes that the fixes must land in a fixed order.

The conversion route starts from a published single-turn task.
It retains that task's repository state, gold patch, test lists and container image, and changes only how the issue is presented across turns.

\subsection{From source to task package}

For a mined cascade, the pipeline retrieves both merged fixes, builds the
grading image and tests, maps acceptance points to the two phases, and creates
one task package and information pool. For a converted task, it preserves the
upstream repository state, image and test verdicts while reorganizing the
user-visible report into turn material. Both routes use the same admission
checks below.

\subsection{Information pools}
\label{app:pools}

Each task has one information-pool file.
The simulator selects from the file before applying a persona's wording.

\begin{center}
\small
\begin{tabular}{@{}lp{9.6cm}@{}}
\toprule
Field & Content \\
\midrule
clear symptom & one reproducible user-visible observation and a broad request \\
vague symptom & the same problem as an impression, without implementation identifiers \\
correct idea & a low-confidence direction consistent with the cause, not a repair step \\
wrong idea & a plausible direction that points somewhere real but is not causal \\
follow-up set & the same four fields for the second issue, phrased as appearing after the first fix \\
phase map & the acceptance points assigned to each phase \\
\bottomrule
\end{tabular}
\end{center}

Because pool construction can inspect the gold patch, it applies two leak
guards. Gold-introduced identifiers are forbidden and checked after
generation. For cascades, shared scope named in both issue titles is removed
unless it is already delivered by the first patch. A task is rejected when
either check cannot produce a clean pool.

\subsection{Admission}
\label{app:admission}

Admission starts the task's grading container and calls the evaluation's test runner in two states.
All four outcomes in Table~\ref{tab:admission} are required.

\begin{table}[h]
\centering
\small
\caption{The four admission outcomes.}
\label{tab:admission}
\begin{tabular}{@{}llll@{}}
\toprule
State & Test set & Required & Establishes \\
\midrule
base & pass-to-pass & pass & existing behavior is intact \\
base & fail-to-pass & fail & the task remains open \\
gold & pass-to-pass & pass & the fix introduces no covered regression \\
gold & fail-to-pass & pass & the task is closeable \\
\bottomrule
\end{tabular}
\end{table}

The verdict records which run failed and carries fingerprints of the grading protocol and task image.
A fingerprint mismatch makes the verdict stale.

For a cascade, an intermediate check additionally requires the first phase to
pass after the first fix while the second remains open and regression guards
stay green.

\paragraph{Agent solvability review.}
Mechanical admission establishes that a task can be graded, but not that its
information pool is sufficient. We therefore applied the same read-only
solvability review to all 72 tasks, comparing the agent-visible material with
the tested behavior. The model review was advisory: a human curator read its
findings, decided the revisions and made the final retention decision. Pools
requiring revision were updated from the review feedback and retained only
after re-review. Gold patches and tests served only as review evidence and were
not exposed to agents.

\section{How the user is simulated}
\label{app:personas}

The simulator combines task content from the information pool with a persona that controls clarity, direction and stubbornness.

\subsection{Persona design}

The persona definitions and utterance policy were designed by the authors.
Production sessions calibrate their frequencies and transitions; they do not
define the policy automatically.

\paragraph{Symptom axis.}
The persona grid uses two ways of reporting the same user-observable problem.
A \emph{clear} symptom names a precise object, reproducible condition or
explicit acceptance target. A \emph{vague} symptom reports a user-facing
impression or broad goal without implementation identifiers.

\paragraph{Cause axis.}
Each symptom setting is crossed with four cause settings.
The cause axis determines what direction the user volunteers and how the user
reacts when challenged. Table~\ref{tab:cause-axis} defines its four settings.

\begin{table}[h]
\centering
\small
\caption{The four settings on the cause axis.
Each is paired with either a clear or vague symptom.}
\label{tab:cause-axis}
\begin{tabular}{@{}p{2.1cm}p{4.8cm}p{5.2cm}@{}}
\toprule
Cause & Opening behavior & Response to disagreement \\
\midrule
None & Gives no diagnosis or repair direction. & Continues with observable symptoms and answers direct questions. \\
Correct & Offers the correct direction as a low-confidence guess. & Gives way when the agent provides a better account. \\
Off-target & Offers a plausible but mistaken direction. & Drops the direction after reasoned disagreement and continues as \emph{none}. \\
Wrong & Opens with the same mistaken direction as \emph{off-target}. & Holds it against reasoning and yields only to test, code or reproduction evidence. \\
\bottomrule
\end{tabular}
\end{table}

The off-target and wrong settings are identical in the opening turn.
They separate only through the agent's later response, which makes their distinction observable only in a multi-turn session.

\subsection{Log labeling and transition chains}
\label{app:transition}

We labeled the opening and later user turns of \nsessions sessions from an
internally deployed commercial coding assistant. Sessions without a code change
were excluded, leaving 13{,}530. A language model assigned the symptom and cause
labels using the conversation and a digest of the eventual findings. It chose
among nine options, the eight cells and \fix{other}, so an opening that sits on
neither axis could be reported as such. The 182 openings labeled \fix{other} are counted in
Table~\ref{tab:persona} and carry no chain, since a persona the scheme does not
name cannot be instantiated in the simulator.
We then read ten sessions per category by hand, and our reading agreed with the
assigned label in each one.
The underlying sessions are proprietary; we release only the aggregate
statistics and calibrated chains.
Table~\ref{tab:persona} gives the opening distribution.
\begin{table}[t]
\centering
\caption{Opening persona distribution over the 13,530 code-changing sessions. The eight cells cover 98.65\% of them; the rest opened as \fix{other}.}
\label{tab:persona}
\small
\begin{tabular}{llrr}
\toprule
Symptom & Stated cause & Sessions & Share \\
\midrule
clear & \fix{none} & 7,168 & 52.98\% \\
clear & \fix{correct} & 2,575 & 19.03\% \\
clear & \fix{off} & 122 & 0.90\% \\
clear & \fix{wrong} & 197 & 1.46\% \\
vague & \fix{none} & 3,103 & 22.93\% \\
vague & \fix{correct} & 123 & 0.91\% \\
vague & \fix{off} & 27 & 0.20\% \\
vague & \fix{wrong} & 33 & 0.24\% \\
\midrule
\multicolumn{2}{l}{Covered by the eight cells} & 13,348 & 98.65\% \\
\multicolumn{2}{l}{\fix{other}} & 182 & 1.35\% \\
\bottomrule
\end{tabular}
\end{table}

\begin{figure}[t]
\centering
\includegraphics[width=\linewidth]{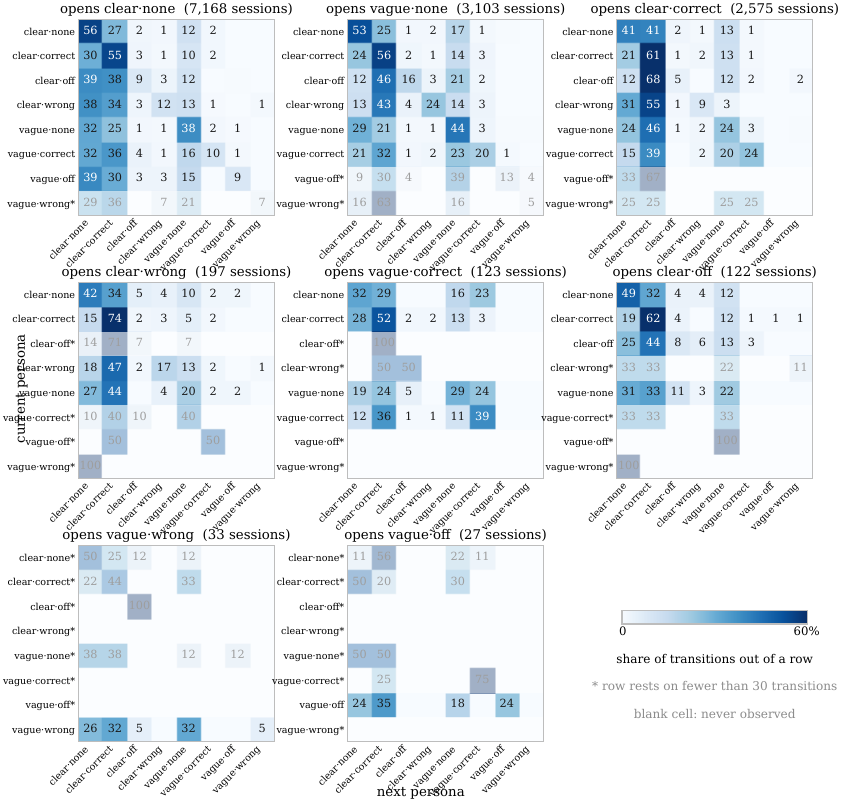}
\caption{The eight turn-to-turn transition matrices, one for each opening
persona. Rows are current personas and columns are next personas.
Rows with fewer than 30 observed transitions are greyed and marked with an asterisk, and entries below 0.5\% are left blank.}
\label{fig:transition}
\end{figure}

The harness draws an opening from the eight cells renormalized to one, then
conditions its transition chain on that opening, yielding
one matrix for each of the eight openings. It drops the end-of-session column,
renormalizes each row and uses the resulting matrix only until the task is
solved. Figure~\ref{fig:transition} shows all eight matrices.

Most transition mass returns to a clear symptom with either no stated cause or a correct cause.
The two mistaken settings are pooled in aggregate reporting because they share an opening and are individually sparse, while the runtime trajectory retains which one is played.

\subsection{Utterance policy}

After selecting guidance, the simulator speaks as a software user rather than
a code reviewer. It states user-visible wants, observations and expectations,
does not expose evaluator or implementation details, answers direct
clarifications first, and realizes the selected symptom clarity and cause
setting.

The release selector reads the agent's response and canonical test results, then applies the fixed priority in Table~\ref{tab:release_priority}.
The tests identify what remains open, but every outward message must stay within the information pool.
Here a core business boundary is behavior required by the primary workflow,
whereas a secondary edge case is remaining acceptance behavior outside that
workflow.

\begin{table}[h]
\centering
\small
\caption{Priority for selecting guidance from the information pool.}
\label{tab:release_priority}
\begin{tabular}{@{}cp{4.4cm}p{7.1cm}@{}}
\toprule
Priority & Trigger & Release \\
\midrule
1 & The agent asks a direct question. & Answer that question. \\
2 & A user-observable core failure remains. & Release the most blocking observable symptom. \\
3 & A core business boundary remains uncovered. & Release the relevant boundary. \\
4 & Only a secondary edge case remains. & Release the relevant edge case. \\
\bottomrule
\end{tabular}
\end{table}

The first applicable category takes priority, and a solved task releases nothing further.

The sampled trajectory is fixed, while the played cause can be weakened by the
interaction rules.
An off-target cause becomes \emph{none} after the agent directly contradicts it
with a code- or behavior-based explanation.
A wrong cause remains until contradicted by an executed test, a reproduction or
an inspected code fact.
A persona with no cause never acquires one, and a correct cause never becomes an implementation instruction.
If the information pool contains no direction corresponding to the sampled
cause, the played cause becomes \emph{none} while symptom clarity is preserved.
The trajectory records both the sampled and played settings.

\subsection{Human validation}
\label{app:people}
\label{app:opening_review}
\label{app:curation}

The evaluation framework can pause before any simulated user turn is sent. At
such a breakpoint a human reads the utterance the simulator is about to send and
checks that a user experiencing the issue could plausibly say it, and that it
stays within the information pool and the sampled persona. A turn must also
react to what the agent has just done while respecting the per-turn information
budget, so the check covers whether it volunteers the cause, the test names or
the shape of the fix. Every task in the released set passed this review at least
once.

When a turn fails the check we do not edit that sentence. We change what
produced it: the generation rule, the utterance-writing prompt or the
information pool, and then regenerate; the regenerated turns go through the
breakpoints again. A single such change applies to every turn generated
afterwards, so the review effort does not grow with the size of the set. Review
at these breakpoints required 50 human-hours in total and no
repository-specific expertise.

\section{Evaluation framework}
\label{app:scoring}

The framework separates environment controls, the Judge, the Reviewer and a post-run audit.

\subsection{Environment controls}

\paragraph{Internal assets.}
The agent sees the agent-facing task statement and its workspace,
but not the source issue or pull-request records, gold patch, fail-to-pass
tests, acceptance list, phase map, scorecard, Judge--Reviewer records or other
evaluation metadata. These assets are mounted outside the workspace, and tool
controls block direct and indirect access to them.

\paragraph{External retrieval.}
General technical documentation is allowed. The agent may not retrieve this
task's issues, pull requests, commits, releases, other source versions or
ready-made fixes from code hosts, package registries, mirrors or caches.
External lookups are recorded. A violation is terminal only when
task-specific protected material is returned and used.

\subsection{Judge}

The framework sends the Judge the canonical user message, agent response,
external-lookup and private-access audits, canonical test window, workspace
fingerprints, previous equivalence ruling, remaining turn budget and recent
user messages.
It returns a candidate state, item-level verdicts, supporting evidence and the
next user message. Table~\ref{tab:judge-states} gives the action associated
with each state.

\begin{table}[h]
\centering
\small
\caption{Judge states.}
\label{tab:judge-states}
\begin{tabular}{@{}clp{4.4cm}p{5.1cm}@{}}
\toprule
State & Name & Meaning & Framework action \\
\midrule
\textbf{U} & unchanged & The agent asks or discusses without changing the workspace. & Return the user's answer and continue; add 0.5 to the weighted turn count. \\
\textbf{F} & continue & At least one acceptance item remains unsatisfied. & Send the selected guidance and begin the next turn. \\
\textbf{C} & framework step & Automated items pass, but another framework step remains. & Run that step and return its evidence to the Judge. \\
\textbf{D} & done & All completion conditions are satisfied. & End the session and record the task as resolved. \\
\textbf{R} & equivalence & A behavior-equivalent test rewrite is requested. & Execute the rewrite through the equivalence gates, then settle again. \\
\textbf{E} & judge error & The Judge fails irrecoverably. & Stop and mark the run failed. \\
\textbf{T} & test error & Canonical tests are unavailable because the environment fails. & Stop and mark the run failed. \\
\textbf{X} & policy violation & Protected or externally retrieved task-specific content is used. & Stop and mark the result unscorable. \\
\bottomrule
\end{tabular}
\end{table}

The framework first handles policy and environment failures, then derives
item-level verdicts from canonical tests. Workspace changes distinguish
\textbf{U} from active attempts; unresolved items yield \textbf{F}, completed
items yield \textbf{D}, and remaining framework work yields \textbf{C}. State
\textbf{R} must carry a candidate test rewrite rather than a verbal equivalence
claim. Evidence remains separate from the user-facing message.

\subsection{Reviewer and settlement}

The Reviewer receives the Judge candidate and the same framework evidence.
It checks each verdict against canonical evidence, the next message against the
persona, release priority and information pool, and each equivalence request
against its formal contract. Table~\ref{tab:reviewer-actions} gives its actions.

\begin{table}[h]
\centering
\small
\caption{Reviewer actions and their framework effects.}
\label{tab:reviewer-actions}
\begin{tabular}{@{}p{2.1cm}p{2.4cm}p{7.7cm}@{}}
\toprule
Context & Action & Framework effect \\
\midrule
Turn review & \textbf{KEEP} & Dispatch the Judge candidate unchanged. \\
Turn review & \textbf{REWRITE} & Replace it with a complete response carrying the revised state, verdicts, evidence and user message. \\
Turn review & \textbf{UNAVAILABLE} & Stop the run rather than dispatch an unreviewed candidate. \\
Equivalence & \textbf{APPROVE} & Allow a mechanically valid rewrite to update the requested items. \\
Equivalence & \textbf{REJECT} & Keep the canonical failure even when the mechanical gates pass. \\
\bottomrule
\end{tabular}
\end{table}

Mechanical settlement then applies four rules: a canonical pass overrides a
model-written N; an equivalence Y without a valid attestation becomes N; a
claimed no-edit state with a changed workspace fingerprint is re-asked with
\textbf{U} disabled; and an acceptance-item set that differs from the task
contract fails closed.

For an equivalence request, the Reviewer makes only the final behavioral
approve-or-reject decision. Algorithm~\ref{alg:equivalence} shows how execution
evidence and that decision are settled.

\begin{algorithm}[h]
\caption{Settle a behavior-equivalent test rewrite.}
\label{alg:equivalence}
\footnotesize
\begin{algorithmic}[1]
\Require candidate workspace $W$, base workspace $W_0$, rewrite request $q$,
regression tests $P$, acceptance scorecard $s$
\Ensure attestation $a$ and updated scorecard $s'$
\If{$W$ has no new change or $q$ contains no executable test rewrite}
  \State \Return rejected, no scorecard update
\EndIf
\State run the rewritten test and $P$ on $W$
\If{the rewritten test fails or any test in $P$ regresses}
  \State \Return rejected at the candidate gate, no scorecard update
\EndIf
\State run the rewritten test on $W_0$
\If{it passes on $W_0$}
  \State \Return rejected at the base gate, no scorecard update
\EndIf
\State ask the Reviewer whether the rewrite preserves the requested behavior
\If{the Reviewer rejects or is unavailable}
  \State \Return rejected at the behavioral gate, no scorecard update
\EndIf
\State $a \gets$ commands, candidate/base outcomes, regression outcomes and Reviewer ruling
\State update only the requested items in $s$ and settle the resulting \textbf{F}, \textbf{C} or \textbf{D} state
\State \Return approved, $a$, updated scorecard $s'$
\end{algorithmic}
\end{algorithm}

A Reviewer failure therefore closes the path rather than dispatching an
unreviewed decision.

\subsection{Audit rubric}
\label{app:rubric}

We maintain the rubric by reading trajectories after each evaluation batch and
adding rules for newly observed failure modes. This requires about one
human-day per batch. An audit agent applies the 13 items in
Table~\ref{tab:audit-rubric}.

\begin{table}[h]
\centering
\footnotesize
\caption{Post-run audit rubric. A flag records a condition for review and does
not by itself make the score untrustworthy.}
\label{tab:audit-rubric}
\begin{tabular}{@{}p{5.0cm}p{7.5cm}@{}}
\toprule
Item & Scope \\
\midrule
\texttt{leak\_test} & Hidden tests, acceptance lists or test-side secrets reached the agent. \\
\texttt{leak\_issue} & The prompt exposed wording or links from the source issue or pull request. \\
\texttt{issue\_allowed} & The word \emph{issue} appeared only in its ordinary product or repository meaning and was harmless. \\
\texttt{judge\_test\_blocked} & The Judge reported that the canonical test command could not be executed. \\
\texttt{test\_bridge\_broken} & The test bridge or environment produced no trustworthy test signal. \\
\texttt{manual\_pytest\_workaround} & The Judge bypassed the canonical runner with an ad hoc test command. \\
\texttt{fabricated\_pass} & The Judge awarded a pass without a trustworthy executed test signal. \\
\texttt{web\_cheat} & An external lookup returned a task-specific answer that the agent used. \\
\texttt{model\_fake\_tests} & The agent fabricated or misreported a test result. \\
\texttt{git\_cheat} & Local repository history exposed a post-base upstream fix that the agent used. \\
\texttt{equivalence\_override} & A canonical failure was replaced by a framework-attested equivalent pass; this is informational. \\
\texttt{reviewer\_misbehavior} & The Reviewer accepted or reversed a decision without evidence, exceeded its role or was unavailable while a decision was dispatched. \\
\texttt{other\_nonstandard} & A concrete non-standard behavior not covered by another item. \\
\bottomrule
\end{tabular}
\end{table}

Trustworthiness is judged separately from rubric membership. A blocked or
detected attempt can be flagged while the score remains trustworthy. A score
is marked untrustworthy only when the flagged behavior changed the final
verdict outside a valid equivalence override.

\section{Details of the experiments}
\label{app:percell}

This appendix records the settings and measurements needed to read the main results.

\subsection{Experimental setup}

Each task receives a persona trajectory before any agent runs.
The trajectory is drawn with a seed determined by the task, opening persona and run index, then frozen in the cohort manifest.
Agents evaluated on the same task receive the same trajectory, while the
session stops as soon as the acceptance points are met.
Single-issue tasks run for at most five turns, and two-issue tasks for at most eight.
The archived run manifest records the provider model identifier and per-run
settings; Table~\ref{tab:systems-scaffolds} gives the model--scaffold assignment
used to read the reported results.

\begin{table}[t]
\centering
\caption{The model and scaffold behind each agent.}
\label{tab:systems-scaffolds}
\small
\begin{tabular}{@{}ll@{}}
\toprule
Model & Scaffold \\
\midrule
Claude Opus 5        & Claude Code \\
Claude Fable 5.1     & Claude Code \\
Grok-4.6             & Claude Code \\
GPT-6-Astra          & Codex \\
DeepSeek-v4.1-Flash  & DSH \\
GPT-5.6-Sol          & Codex \\
GLM-5.3              & Z.ai Code \\
Gemini-3.8-Flash     & Claude Code \\
Hunyuan-4.0-preview  & Claude Code \\
Doubao-Seed-2.1-Pro  & Claude Code \\
Kimi-K3              & Kimi Code \\
\bottomrule
\end{tabular}
\end{table}

The judging roles are balanced between two model families.
In one assignment Claude Code--Opus 5 judges and GPT-5.6-Sol on the Codex scaffold reviews, and in the other the roles are reversed.
The Reviewer sees the Judge's candidate response and the framework evidence, not a new execution of the agent.

\subsection{Confidence intervals and paired comparisons}
\label{app:stats}

The sampling unit is the task.
All eleven agents run the same 72 tasks, so every comparison between two agents is paired within the task and the spread of task difficulty drops out of it.
That spread is the larger of the two sources of variation, since the standard deviation of mean LDPR is 0.283 across tasks and 0.073 across agents.
Pairing removes it, which is why every agent is given the same tasks and the same persona trajectories.
Table~\ref{tab:leaderboardci} gives each agent's LDPR with a 95\% interval obtained by resampling tasks with replacement.

\begin{table}[h]
\centering
\caption{LDPR with a 95\% interval from resampling tasks with replacement.}
\label{tab:leaderboardci}
\small
\begin{tabular}{@{}lcc@{}}
\toprule
Model & LDPR & 95\% interval \\
\midrule
Claude Opus 5 & 0.640 & [0.553, 0.722] \\
Claude Fable 5.1 & 0.593 & [0.499, 0.684] \\
Grok-4.6 & 0.569 & [0.486, 0.649] \\
GPT-6-Astra & 0.561 & [0.481, 0.639] \\
DeepSeek-v4.1-Flash & 0.537 & [0.450, 0.621] \\
GPT-5.6-Sol & 0.500 & [0.412, 0.587] \\
GLM-5.3 & 0.496 & [0.403, 0.587] \\
Gemini-3.8-Flash & 0.458 & [0.366, 0.551] \\
Hunyuan-4.0-preview & 0.457 & [0.369, 0.545] \\
Doubao-Seed-2.1-Pro & 0.456 & [0.370, 0.542] \\
Kimi-K3 & 0.391 & [0.301, 0.485] \\
\bottomrule
\end{tabular}
\end{table}

Comparisons between agents use a paired permutation test on the 72 per-task differences, with the 55 pairs corrected by Benjamini-Hochberg at a 5\% false discovery rate.
What the evaluation measures matters as much as how many turns it allows.
On the same tasks and the same 55 pairs, LDPR separates 17 pairs, against 5 for whether a task is resolved on the first turn and 5 for whether it is resolved at all.
Resolution over the full budget carries the highest base rate of the three at 71.8\% and still separates the fewest pairs, so the discount for length is what carries the additional signal.

\subsection{Cost}

Table~\ref{tab:percell_cost} separates the agent from the judging side.

\begin{table}[t]
\centering
\caption{Per-task cost, separated between the agent and judging.
Tokens are in thousands and exclude cache reads; costs are in CNY. Metrics use
available telemetry: agent columns are means, while judging costs and edit
sizes are medians.}
\label{tab:percell_cost}
\small
\setlength{\tabcolsep}{4pt}
\begin{tabular}{@{}l|ccc|ccc|ccc@{}}
\toprule
& \multicolumn{3}{c|}{Agent} & \multicolumn{3}{c|}{Judging side} & & \multicolumn{2}{c}{Edit size} \\
\cmidrule(lr){2-4}\cmidrule(lr){5-7}\cmidrule(l){9-10}
Model & Requests & Tokens & Cost & Requests & Judge & Reviewer & Wall & Files & Lines \\
\midrule
Claude Opus 5 & 178 & 22279 & 170.4 & 71 & 33.5 & 30.7 & 128 & 5 & 572 \\
Claude Fable 5.1 & 273 & 38635 & 710.5 & 75 & 46.4 & 34.8 & 675 & 6 & 633 \\
Grok-4.6 & 177 & 18175 & 110.8 & 82 & 40.4 & 37.1 & 168 & 6 & 696 \\
GPT-6-Astra & 158 & 13725 & 264.2 & 114 & 61.9 & 38.8 & 805 & 7 & 723 \\
DeepSeek-v4.1-Flash & 386 & 28355 & 4.3 & 134 & 72.3 & 30.4 & 278 & 3 & 184 \\
GPT-5.6-Sol & 524 & 54633 & 279.2 & 70 & 36.6 & 34.8 & 200 & 3 & 221 \\
GLM-5.3 & 243 & 51318 & 70.6 & 98 & 54.2 & 37.3 & 162 & 6 & 1146 \\
Gemini-3.8-Flash & 748 & 68865 & 107.3 & 197 & 85.9 & 45.3 & 746 & 4 & 283 \\
Hunyuan-4.0-preview & 197 & 16501 & 20.8 & 113 & 57.9 & 47.8 & 270 & 4 & 250 \\
Doubao-Seed-2.1-Pro & 307 & 28466 & 50.6 & 32 & 37.5 & 35.9 & 422 & 4 & 314 \\
Kimi-K3 & 146 & 13950 & 37.3 & 80 & 48.8 & 45.5 & 103 & 4 & 390 \\
\bottomrule
\end{tabular}
\end{table}

\subsection{Reviewer rewrites by model}
\label{app:reviewer-models}

Table~\ref{tab:reviewerbymodel} expands the aggregate Reviewer analysis in
Section~\ref{sec:val:reviewer} across the eleven leaderboard models.
The rewrite rate is calculated over \texttt{turn\_decision} records within each
model--reviewer configuration; finalize confirmations and other audit
checkpoints are excluded.
Each row's objection columns are shares of the objections that this Reviewer
filed for this model, so the four sum to 100\% along every row and the two
adjacent rows can be read independently.

\begin{table}[h]
\centering
\caption{Reviewer rewrites by leaderboard model and Reviewer. Rewrite is the
share of reviewed turns rewritten; the remaining columns break down that row's
own objections and sum to 100\% across each row.}
\label{tab:reviewerbymodel}
\small
\setlength{\tabcolsep}{4pt}
\renewcommand{\arraystretch}{1.04}
\begin{tabular*}{\linewidth}{@{\extracolsep{\fill}}llrrrrr@{}}
\toprule
Model & Reviewer & Rewrite & Evidence & Guidance & Equiv. & Persona \\
\midrule
Claude Opus 5 & Claude Code & 30.0\% & 32.6\% & 44.0\% & 9.3\% & 14.1\% \\
 & Codex & 64.8\% & 70.9\% & 13.9\% & 7.1\% & 8.1\% \\
\cmidrule(lr){1-7}
Claude Fable 5.1 & Claude Code & 24.2\% & 48.1\% & 30.1\% & 7.9\% & 13.9\% \\
 & Codex & 70.5\% & 66.3\% & 17.4\% & 11.7\% & 4.6\% \\
\cmidrule(lr){1-7}
Grok-4.6 & Claude Code & 21.9\% & 40.8\% & 38.5\% & 12.9\% & 7.8\% \\
 & Codex & 65.3\% & 78.8\% & 12.7\% & 6.4\% & 2.1\% \\
\cmidrule(lr){1-7}
GPT-6-Astra & Claude Code & 25.9\% & 44.5\% & 39.9\% & 4.6\% & 11.0\% \\
 & Codex & 69.4\% & 65.3\% & 14.9\% & 11.9\% & 7.9\% \\
\cmidrule(lr){1-7}
DeepSeek-v4.1-Flash & Claude Code & 31.7\% & 54.0\% & 41.2\% & 0.0\% & 4.8\% \\
 & Codex & 74.4\% & 68.5\% & 18.0\% & 7.9\% & 5.6\% \\
\cmidrule(lr){1-7}
GPT-5.6-Sol & Claude Code & 25.7\% & 41.4\% & 46.4\% & 4.9\% & 7.3\% \\
 & Codex & 45.3\% & 69.3\% & 12.2\% & 13.9\% & 4.6\% \\
\cmidrule(lr){1-7}
GLM-5.3 & Claude Code & 26.5\% & 32.3\% & 51.7\% & 1.5\% & 14.5\% \\
 & Codex & 66.4\% & 74.4\% & 12.4\% & 8.6\% & 4.6\% \\
\cmidrule(lr){1-7}
Gemini-3.8-Flash & Claude Code & 25.9\% & 39.7\% & 43.3\% & 11.2\% & 5.8\% \\
 & Codex & 70.9\% & 75.4\% & 8.7\% & 8.7\% & 7.2\% \\
\cmidrule(lr){1-7}
Hunyuan-4.0-preview & Claude Code & 27.8\% & 41.9\% & 40.5\% & 3.7\% & 13.9\% \\
 & Codex & 56.6\% & 62.2\% & 24.2\% & 6.2\% & 7.4\% \\
\cmidrule(lr){1-7}
Doubao-Seed-2.1-Pro & Claude Code & 30.1\% & 50.0\% & 34.1\% & 5.3\% & 10.6\% \\
 & Codex & 48.5\% & 72.2\% & 16.4\% & 6.3\% & 5.1\% \\
\cmidrule(lr){1-7}
Kimi-K3 & Claude Code & 15.2\% & 34.0\% & 46.9\% & 2.1\% & 17.0\% \\
 & Codex & 59.8\% & 67.6\% & 11.5\% & 14.2\% & 6.7\% \\
\bottomrule
\end{tabular*}
\end{table}

\clearpage
\subsection{Equivalence gates}
\label{app:artifacts}

Table~\ref{tab:equivalence-gates} assigns the 62 refused equivalence requests
to the first gate each fails.
The rewritten test must pass on the candidate workspace without regressing a pass-to-pass test.
It must still fail at the base state, or it no longer tests the proposed fix.
If both mechanical gates hold, the Reviewer can still reject a rewrite that drops behavior required by the issue.

\begin{table}[h]
\centering
\small
\caption{First failed gate among refused equivalence requests.}
\label{tab:equivalence-gates}
\begin{tabular}{@{}p{2.8cm}p{7.1cm}r@{}}
\toprule
First failed gate & Condition & Requests \\
\midrule
Candidate execution & Rewritten acceptance tests fail, or a pass-to-pass test regresses, on the candidate workspace. & 47 \\
Base discrimination & The rewritten test also passes on the unmodified base workspace. & 6 \\
Behavioral review & Both execution gates pass, but the rewrite drops behavior required by the issue. & 9 \\
\bottomrule
\end{tabular}
\end{table}

\section{Case studies}
\label{app:cases}

The cases below follow one mined cascade from its source issues to a graded
session, then illustrate a full session, equivalence decisions and a Reviewer
rewrite.

\subsection{From two issues to a cascade task}
\label{app:case_cascade}

\texttt{apm-casc-2376-2377} is built from two merged issues in
\texttt{microsoft/apm}, \emph{APM generates npx launcher for Docker-runtime MCP
package} (\#2376) and \emph{\texttt{-{}-target vscode} drops custom Docker
arguments for MCP server} (\#2377). They were paired because the second pull
request declares that it follows the first. The code forces the same order,
since the launcher must be routed to the Docker branch before any Docker
argument exists to be dropped. One of the second-phase acceptance points is
named \texttt{test\_launcher\_is\_no\_longer\_the\_bare\_fallback}, and it is
only reachable once the first fix lands. The separability check confirms this
mechanically, with the second phase still failing when the first fix is applied
alone.

The task package grades two acceptance points per phase and holds three
regression tests. The information pool carries a clear and a vague statement of
each phase's symptom together with a correct and an incorrect candidate cause,
from which the sampled persona selects. Table~\ref{tab:casccase} shows what the
agent receives at each phase. Note how little the task statement
gives away. The agent sees the first issue's title and a request to locate and
fix the cause in the current repository, with no test names, no second issue and
no gold patch.

\begin{table}[h]
\centering
\small
\caption{The two phases of \texttt{apm-casc-2376-2377}. The user turns are the
frozen utterances for a clear-symptom, correct-cause persona, translated from
the deployment language. Both were approved at the breakpoint review.}
\label{tab:casccase}
\begin{tabular}{@{}p{1.5cm}p{5.6cm}p{5.6cm}@{}}
\toprule
& Phase 1 (\#2376) & Phase 2 (\#2377) \\
\midrule
User turn &
Generating a launcher for a Docker-runtime package produces \texttt{npx}
instead, and it reports the package cannot be found. I suspect the runtime type
is not recognized and the package falls through to the generic launcher branch. &
For the same Docker package, generating a configuration with
\texttt{-{}-target vscode} makes my custom Docker options disappear. I suspect
the per-target conversion loses the Docker arguments. \\
\addlinespace
Acceptance &
Registry name maps to Docker, and the target generates a Docker launcher
carrying the image. &
Value entries become \texttt{docker run} options, and the launcher is no longer
the bare fallback. \\
\addlinespace
Revealed &
At the opening. &
Only after both first-phase points pass. \\
\bottomrule
\end{tabular}
\end{table}

Because the second issue is withheld until both first-phase points pass, the
task cannot be closed in a single turn regardless of how strong the agent is.

\subsection{A full session, end to end}
\label{app:transcript}

\texttt{add-date-provider-pattern-arg} asks for a way to serialize a generated date object.
Table~\ref{tab:sessioncase} follows the three turns of the session.

\begin{table}[h]
\centering
\small
\caption{The three turns of \texttt{add-date-provider-pattern-arg}. Tests run
after every turn, and what the simulated user says next is selected from the
information pool against that result.}
\label{tab:sessioncase}
\begin{tabular}{@{}p{0.8cm}p{3.6cm}p{4.2cm}p{3.9cm}@{}}
\toprule
Turn & User turn & Agent response & Tests after the turn \\
\midrule
1 &
The generated date object cannot be dumped to JSON, and an option returning a string would help. &
Adds \texttt{as\_string}, which returns a fixed ISO string. &
The regression guard stays green and the feature acceptance points stay red. \\
\addlinespace
2 &
The output format has to be selectable. &
Lets \texttt{as\_string} take a format string. &
Unchanged, with the feature acceptance points still red. \\
\addlinespace
3 &
Calls that pass \texttt{pattern} are still rejected. &
Adds \texttt{pattern} to the affected date APIs and propagates it through the call chain, keeping the object return when it is omitted. &
Every feature acceptance point turns green and the regression guard stays green. \\
\bottomrule
\end{tabular}
\end{table}

After the third turn the Judge returned \textbf{D} and the cross-family Reviewer kept that decision.
Both of the first two implementations answer what had been said so far, and neither reaches the argument name that the repository already exposes, which the session only surfaces on the third turn.

\subsection{An equivalent pass, granted and refused}
\label{app:case_equiv}

Table~\ref{tab:equivcases} places one grant beside one refusal.
Both requests reached the mechanical gates, so the difference is what the rewritten test continued to require.
The granted request is from \texttt{add-task-timeout-cli-option}, and the refused request is from \texttt{support-path-typed-lists-cli}.

\begin{table}[h]
\centering
\small
\caption{Two equivalence requests and the behavior retained by their rewritten tests.}
\label{tab:equivcases}
\begin{tabular}{@{}p{2.1cm}p{4.9cm}p{5.2cm}@{}}
\toprule
Request & Rewrite & Ruling \\
\midrule
Granted &
Replace a literal comparison against the task dictionary with a non-default command-line timeout, load the resulting play, and inspect the effective timeout on its task object. &
Granted.
The base state still failed, the candidate and regression tests passed, and the Reviewer found the command-line-to-task behavior equivalent.
The Reviewer separately noted dropped assertions on asynchronous fields and an unrelated wording edit. \\
\addlinespace
Refused &
Drive the command through \texttt{sys.argv} and capture values received by a wrapped function instead of checking the public \texttt{parse\_args(args)} and \texttt{run()} return contracts. &
Refused.
The mechanical gates passed, but the rewrite removed behaviors named by the contract, introduced mocks the contract excluded, and replaced an optional list of paths with a scalar path. \\
\bottomrule
\end{tabular}
\end{table}

The refusal shows why a candidate passing a rewritten test is not sufficient.
The Reviewer checks whether the rewrite preserved the behavior being graded, even after the base-fail and candidate-pass gates hold.

\subsection{A reviewer rewrite}
\label{app:case_rewrite}

On turn one of \texttt{strip-author-name-honorifics} the Judge left the failed items at \textbf{N} and decided to continue for another turn, and the Reviewer kept both.
What the Reviewer changed was the evidence behind that verdict and the message the simulated user was about to send.
Table~\ref{tab:rewritecase} separates the two, which are the first two of the four review domains in Table~\ref{tab:reviewercat}.

\begin{table}[h]
\centering
\small
\caption{The two objections filed on turn one of
\texttt{strip-author-name-honorifics} and how the Reviewer settled each.}
\label{tab:rewritecase}
\begin{tabular}{@{}p{1.6cm}p{5.3cm}p{5.3cm}@{}}
\toprule
Domain & What the Judge wrote & How the Reviewer settled it \\
\midrule
Evidence &
A zero return code together with a collection-time import error, neither of which the canonical test window shows. &
Recorded what that window holds, a non-zero return code with the fail-to-pass set failing and the pass-to-pass set passing, and anchored the diagnosis to the workspace changes. \\
\addlinespace
Guidance &
A next message carrying evaluator detail, which the Judge is required to keep separate from its evidence, and a candidate state the framework could not read. &
Restored the state to where the framework reads it and cut the next message to one user-visible observation, that the saved name still carried \texttt{Mr.} \\
\bottomrule
\end{tabular}
\end{table}

The verdict is the same before and after, so a rewrite of this kind does not move the score of the run it lands in.
What it changes is the record the next turn is built on and the utterance the agent receives.

\section{Opening clarity across the leaderboard}
\label{app:clarity}

Every task carries a fixed opening persona, shared by all agents evaluated on
it, so the leaderboard splits into a group whose opening states the symptom
precisely and a group that states it as an impression. Table~\ref{tab:clarity}
reports LDPR on each group. This is an observational split, and the two groups
hold different tasks, so the level of either group carries task difficulty along
with the effect of how the request is phrased. What the split does isolate is
how the gap varies between agents, since the two groups hold the same tasks for
every agent and task difficulty therefore cancels in that comparison. The gap
runs from 0.268 for Claude Fable 5.1 down to $-$0.088 for DeepSeek-v4.1-Flash,
a spread of 0.356 measured across all eleven agents.

\begin{table}[t]
\centering
\caption{LDPR by the clarity of the opening symptom, over the 55 tasks whose
opening states the symptom precisely and the 17 that state it as an impression.
$\Delta$ is the difference between the two groups, with its standard error.}
\label{tab:clarity}
\small
\setlength{\tabcolsep}{4pt}
\begin{tabular*}{\linewidth}{@{\extracolsep{\fill}}lccc@{}}
\toprule
Model & Clear & Vague & $\Delta$ \\
\midrule
Claude Opus 5       & 0.671 & 0.540 & +.131\,$\pm$\,.106 \\
Claude Fable 5.1    & 0.656 & 0.388 & +.268\,$\pm$\,.123 \\
Grok-4.6            & 0.599 & 0.474 & +.125\,$\pm$\,.101 \\
GPT-6-Astra         & 0.603 & 0.423 & +.181\,$\pm$\,.102 \\
DeepSeek-v4.1-Flash & 0.516 & 0.604 & $-$.088\,$\pm$\,.109 \\
GPT-5.6-Sol         & 0.549 & 0.341 & +.208\,$\pm$\,.101 \\
GLM-5.3             & 0.526 & 0.399 & +.128\,$\pm$\,.104 \\
Gemini-3.8-Flash    & 0.502 & 0.318 & +.184\,$\pm$\,.111 \\
Hunyuan-4.0-preview & 0.500 & 0.315 & +.185\,$\pm$\,.093 \\
Doubao-Seed-2.1-Pro & 0.495 & 0.329 & +.165\,$\pm$\,.103 \\
Kimi-K3             & 0.411 & 0.327 & +.084\,$\pm$\,.101 \\
\bottomrule
\end{tabular*}
\end{table}

\end{document}